\documentclass[10pt,twocolumn,letterpaper]{article}

\usepackage{cvpr}              

\usepackage{graphicx}
\usepackage{amsmath}
\usepackage{amssymb}
\usepackage{booktabs}
\usepackage{algorithm}
\usepackage{algpseudocode}

\usepackage{cite}
\usepackage{amsmath,amssymb,amsfonts}
\usepackage{graphicx}
\usepackage{textcomp}
\usepackage{xcolor}
\def\BibTeX{{\rm B\kern-.05em{\sc i\kern-.025em b}\kern-.08em
    T\kern-.1667em\lower.7ex\hbox{E}\kern-.125emX}}
\usepackage{booktabs}
\usepackage{algorithm}
\usepackage{algpseudocode}
\usepackage{pgfplots}
\usepackage{tabularx}
\usepackage{cite}
\usepackage[top=1.0in, bottom=1.1in, inner=0.75in, outer=0.75in]{geometry}
\usepackage{multirow}
\usepackage[pagebackref,breaklinks,colorlinks]{hyperref}

\usepackage[capitalize]{cleveref}
\crefname{section}{Sec.}{Secs.}
\Crefname{section}{Section}{Sections}
\Crefname{table}{Table}{Tables}
\crefname{table}{Tab.}{Tabs.}

\def\cvprPaperID{*****} 
\def\confName{CVPR}
\def\confYear{2022}

\begin{document}

\title{Efficient Mitigation of Bus Bunching through Setter-Based Curriculum Learning}

\author{Avidan Shah, Danny Tran, Yuhan Tang\\
University of California, Berkeley\\
{\tt\small \{amcshah, dannyltran, yhtang\}@berkeley.edu}
}
\maketitle

\begin{abstract}
   Curriculum learning has been growing in the domain of reinforcement learning as a method of improving training efficiency for various tasks. It involves modifying the difficulty (lessons) of the environment as the agent learns, in order to encourage more optimal agent behavior and higher reward states. However, most curriculum learning methods currently involve discrete transitions of the curriculum or predefined steps by the programmer or using automatic curriculum learning on only a small subset training such as only on an adversary. In this paper, we propose a novel approach to curriculum learning that uses a Setter Model to automatically generate an action space (S), adversary strength $\alpha$, intialization, and bunching strength ($\beta$). Our setter model architecture is a MLP neural network with loss function defined by the previously mentioned three variables. To perform gradient descent, we use the summed log probabilities of the distributions over the three model parameters, $S, \alpha, \text{ and } \beta$, and weight each by negative average reward over the last three timesteps. To incorporate the setter model's inputs into the training of the RL agent, we use Algorithm \ref{tab:sb_curr}, which essentially updates the environment after a certain number of rollouts, then uses the agent's actions to train the setter. The outputs of the setter are then used once again to update the environment.

   We evaluate the efficacy of our automated curriculum learning method in a custom bus-system environment we create, intending to simulate real-world public transportation systems. We adapt our method to work well with the bus-system environment. We compare our novel curriculum learning setter framework to a baseline model with no curriculum learning, as well as two other popular curriculum learning methods frameworks. The results show that curriculum learning with the setter model in its current architecture is slightly outperformed by the baseline model with no curriculum learning. Despite this, it tends to reach maximum rewards faster, is more stability, and considerably easier to use out-of-the-box (due to not needing to develop an actual curriculum) than the other baseline curriculum learning style approaches,``stagnation" and ``budget", meaning it still has potential for future development. 
   
   We also analyze the curriculum of lessons that our setter model creates. We find that when analyzing the action space S that the setter curriculum develops (as seen in Figure \ref{fig:curr_act}) starts with a high exploratory beginning that explores all kinds of different action spaces but around $15000$ steps begins to converge to $S=6$. However, even after primarily converging the model still tries to set various action spaces as the reward plateaus which hints that the setter model tries to break out of the plateau using the action space. A similar pattern is seen when analyzing Figure \ref{fig:curr_adv} which is the adversary strength as the model initially explores different $\alpha$ values but converges to $\alpha=0$ at around $5000$ steps. Again, similarly when analyzing Figure \ref{fig:curr_bun} which plots the bunching strength $\beta$ as training goes on. The similar pattern of a lot of exploration in the beginning is true but eventually converges to a middle bunching strength of $\beta=5$. There is not as many different $\beta$ explorations after converging when compared to the action space $S$ which could indicate that the action space plays a larger role in the curriculum.
   
   In our experiments, we also perform ablation studies on the environment in addition to values that the setter model is able to change. Specifically, we run experiments allowing the setter to modify only one of the action space, perturbation strength, and bunching strength, pairs of the variables, and then all three. The idea behind this is to see which parameters are most crucial to curriculum learning's impact on agent performance. We find that allowing the setter to affect just one variable at a time leads to the best performance, while the experiment run with all three variables is the slowest to converge to maximum reward. This could indicate that the model is being overloaded and does not have the capacity for all $3$, and that a more complex model architecture is required. We also run an ablation study on our baseline model without curriculum learning to test the impact of domain randomization on performance. We note that the agent's performance stagnates without domain randomization, indicating that the introduction of random training examples significantly to model real world cases helps the agent learn.
   
   \color{black}
\end{abstract}

\section{Introduction}
\label{sec:intro}

Transportation and traffic optimization is a well known area of study, especially for reinforcement learning based solutions. We specifically look at the bus bunching problem for the context of this study. The main idea of the problem is to minimize the delays caused by inefficient bus timings for passengers arriving and departing from a system of buses. While the heavy exploration in the area makes innovation and improvement with regards to performance marginal, it simultaneously provides an effective baseline for developing new generalized techniques. Our group is particularly interested in examining curriculum learning and its effect on training efficiency and overall performance. We decide to try a lesser known approach to curriculum learning, in which the curriculum is not fixed or discretely thresholded. Our method for automated curriculum learning involves a curriculum that is dynamically chosen and learned by an adversary network made to increase the difficulty of the agent's training, and defined by multiple forms of input. Our results are shown in the following sections of this paper.

\section{Related Works}
\label{sec:rw}

\paragraph{Bus Systems} Transportation is a commonly studied environment for reinforcement learning experiments, and the bus delay problem in particular is used for a multitude of reasons. The first is likely due to the solution for bus bunching requiring sequential decision making, making it a natural fit for a reinforcement learning agent. If a solution for delays is implemented, buses need to decide which action they will take in between each stop, and this is easy represented in a simulated environment. Additionally, real life traffic systems are dynamic and often times completely random due to accidents, weather, etc., allowing RL models to learn how to adjust to perturbations and not overfit to an unrealistic environment. Many solutions are offered each year to reduce bus waiting time without increasing traffic, such as the one proposed by \cite{shen2023}. We believe that due to the high volume of research towards this problem, it will serve as an excellent baseline for developing new techniques, which if successful can then be applied towards more novel environments.


\paragraph{Curriculum Learning} Curriculum learning is a popular and effective method for improving RL agent performance and efficiency in a variety of environments by guiding agent actions. Several studies have investigated the impact of task sequencing on the learning performance and sample efficiency of RL agents. In general, many propose a curriculum learning framework that intelligently orders tasks to gradually increase in difficulty. By strategically sequencing tasks, agents were shown to not only achieve faster convergence and higher reward states, but improved generalization across a range of environments. Traffic and transportation optimization is also a commonly examined environment for evaluating all types of reinforcement learning algorithms. Other attempts, such as the ones proposed by \cite{narvekar2020curriculum, DBLP:journals/corr/abs-1909-12892}, use curriculum learning to improve upon learning efficiency in various frameworks, however not explicitly in the bus bunching setting. Additionally, others have examined curriculum learning discretely through the lens of an ``adversary," but use a deterministic algorithm rather than a model to define the adversary's actions throughout training. Our research aims to automate a dynamic adversarial alongside action space and initializations in the bus bunching environment through a neural network. One area that we focus on specifically is the idea of a \textit{teacher-student model}, as described by \cite{soviany2022curriculum}. The main idea of this curriculum learning framework is to have one primary model, also referred to as the \textit{student,} learn the actual task, while a \textit{teacher} model learns the optimal learning parameters for the student.


\paragraph{Automated Curriculum Learning} Typically, self play in reinforcement learning has only been possible in symmetric games between two agents. Curriculum learning methods would have to rely on human designed curricula, which often required extensive domain knowledge and time spent. Several methods for automating the creation of a curriculum include varying environment, as suggested by \cite{wangexplor}, in which they trained environments separately with agents, although this resulted in overfitted agents only specialized for their sub-environment. Another technique that has been tried is optimal task selection, as proposed by \cite{baranes}, although this was proven to be extremely computationally expensive once environments grew more complex. However \cite{DBLP:journals/corr/abs-1909-12892} introduce a novel method for the automation of curriculum design for curriculum-based RL. Their introduction of a \textit{setter-solver} paradigm proved to be highly effective in making curriculum learning based methods much more efficient. We aim to build upon this by using a deep neural network that can impact various parts of the training environment.


\section{Method}
\label{sec:method}

\subsection{Bus Environment}
\label{sec:bus_env_1}
\begin{figure}[htbp]
    \centering
    \includegraphics[scale=0.35]{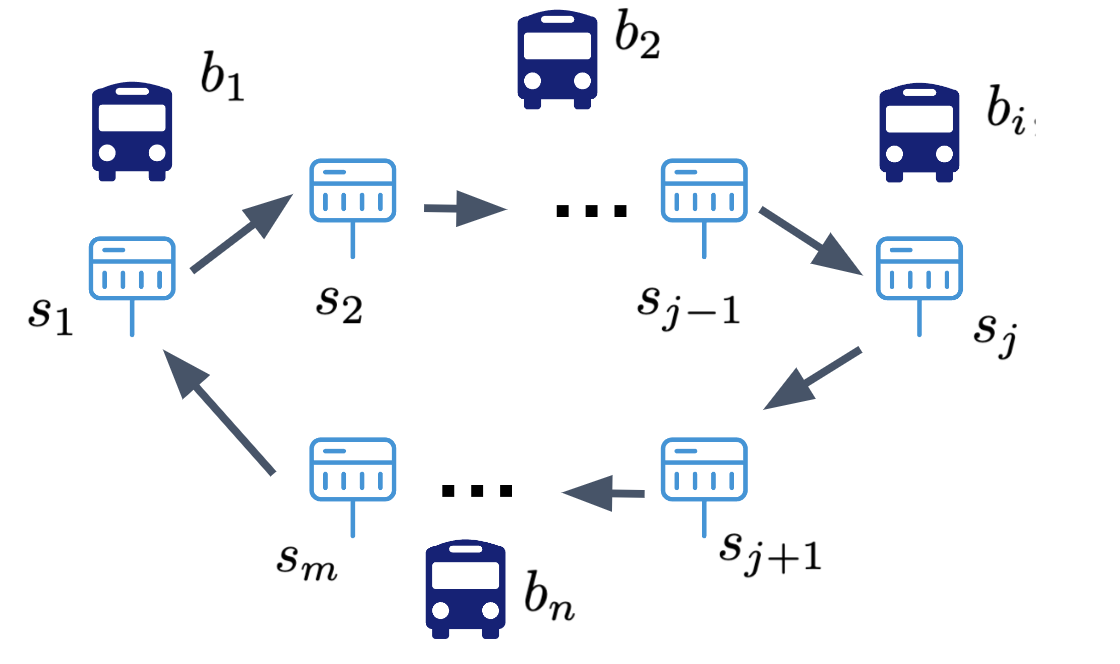}
    \caption{Visualization of the Bus Line simulator}
    \label{fig:bus_line_viz}
\end{figure}

We formulate the bus line corridor model with $m$ uniformly distributed stations, $s_{j}, j = 1,2, ..., m$. There are $n$ buses, $b_{i}, i = 1,2, ..., n$, driving on this route with a constant average speed $v$. The general setup of the framework is shown in Figure \ref{fig:bus_line_viz}. The environment assumes the following assumptions:
\begin{itemize}
    \item The route is a simple loop with stations evenly distributed along \cite{wang_dynamic_2020}.
    \item Without loss of generality, all delays incur at bus stops. Each bus travels at a constant speed, and traffic or other factors do not affect it.
    \item Each bus spends a fixed amount of time at each station for boarding and alighting.
    \item The bus system is a frequency-based system. All the buses are initialized at station $s_{j}$ with a constant interval of headway $H$.
    \item When there is bus bunching at station $s_{j}$, the probability for the passengers to board each bus is equal.
    \item Boarding and alighting occur concurrently and adhere to a proportional pattern based on passenger counts. Each passenger's time to board and exit is consistently $\tau_b$ and $\tau_a$ minutes at every stop. The bus must wait for both processes to complete before it can depart, with the longer of the two processes dictating the overall stop duration at a certain station.
    \item A bus can choose to skip at any station. If a bus decides at station $s_{j}$ to skip the next station $s_{j+1}$, it must alight all the passengers whose destinations are $s_{j+1}$ at station $s_{j}$ and make the next stop at station $s_{j+2}$.
    \item A bus can choose to turn around at any station. Without loss of generality, a bus decides at station $s_{j}, j \leq \lfloor {m}/2 \rfloor $ to turn around. It must alight all the passengers whose destinations are $s_{k}, k=j+1, ..., \lfloor {m}/2 \rfloor$. The symbol \( \lfloor \cdot \rfloor \) denotes the floor function, which rounds down to the nearest integer.
    \item At any given bus stop \( s_j \), the arrival of passengers is described by a Poisson distribution with a  rate of \( \lambda_j \). For a passenger at stop \( s_j \), their destination is selected randomly and with equal chance from the next \( \lfloor {(m-j)}/2 \rfloor \) stops. 
    \item There is a capacity $C$ for each bus. In this paper, $C=60$.
\end{itemize}

In our training environment, 5 passenger statuses are defined as shown in Table \ref{tab:pas_status}. Passengers might be waiting at a station, get on a bus and alighted. When alighted, he might still be waiting since the passenger has not arrived or he might arrive and leave the system. Also, when waiting time is larger than the headway, passengers will have a 50\% possibility to leave the station as a penalty for a long wait.

\begin{table}
\centering
\resizebox{0.65\linewidth}{!}{%
\begin{tabular}{cc}
\hline
Status & Meaning                   \\ \hline
0      & Waiting                   \\ \hline
1      & On bus                    \\ \hline
2      & Alighted and Waiting      \\ \hline
3      & Alighted and Arriving     \\ \hline
4      & Leaving after a long wait \\ \hline
\end{tabular}}
\caption{The 5 different passenger statuses in our Bus Environment. }
\label{tab:pas_status}
\end{table}

The total reward is defined as
$R =\alpha \frac{T_w}{N_w}+\beta \frac{T_b}{N_b}$, where ${T_w}$ refers to the total passenger waiting time, ${T_b}$ refers to the total passenger time on bus, $N_w$ refers to the total number of waiting passengers, $N_b$ refers to the total number of passengers on bus, and $\alpha, \beta\in \mathbb{R}$ are the weighting parameters. 

Both discrete and combined action spaces are feasible for bus operation. On continuous action spaces, standard PPO is unstable when rewards vanish outside bounded support \cite{schulman2017proximal}. Since combined action spaces include continuous action spaces, we choose discrete action spaces for RL in our model. The action values are shown in Table \ref{tab:action_value} and action spaces are shown in Table \ref{tab:action_space}. The action space for holding is discretized into an interval of 10s (For example, an action value of 5 means an action of holding for 50s.) Discrete action space introduces simplicity to the problem-solving process by transforming a combined action space into a manageable set of discrete options for the agent to select from, thereby making the learning algorithm more tractable. Furthermore, combined action spaces are susceptible to stability challenges during training, since minimal alterations in policy could precipitate substantial changes in action outcomes; discretization mitigates this issue, enhancing stability. Moreover, discretization of the action space can significantly reduce the computational resources required for policy evaluation and improvement, which is beneficial for real-world applications where quick decisions may be necessary.

\begin{table*}[]
\centering
\label{Action Value}
\scalebox{1.0}{
\begin{tabular}{cccc}
\hline
\multirow{2}{*}{}       & \multicolumn{2}{c}{\textbf{Combined Action Space}}              & \textbf{Discrete Action Space}    \\ \cline{2-4} 
                        & \textbf{High-level Action} & \textbf{Low-level Action} & \textbf{Discrete Action} \\ \hline
\textbf{Holding} & 0 & Holding time & \begin{tabular}[c]{@{}c@{}}0, 1, 2, 3, 4, 5, 6,\\  7, 8, 9, 10, 11\end{tabular} \\ \hline
\textbf{Skipping}       & 1                          & /                         & 12                       \\ \hline
\textbf{Turning Around} & 2                          & /                         & 13                       \\ \hline
\end{tabular}
}
\caption{Action Values}
\label{tab:action_value}
\end{table*}

\begin{table*}[]
\centering
\label{Action Space}
\begin{tabular}{cccc}
\hline
\multirow{2}{*}{}   & \multicolumn{2}{c}{\textbf{Combined Action Space}}                                               & \textbf{Discrete Action Space} \\ \cline{2-4} 
               & \textbf{High-level Action} & \textbf{Low-level Action} & \textbf{Discrete Action} \\ \hline
\textbf{Holding}             & Discrete(2) & \begin{tabular}[c]{@{}c@{}}Box(low=0, \\ high=1, shape=(1,))\end{tabular} & Discrete(12)          \\ \hline
\textbf{Skipping}       & Discrete(2)       & /                & Discrete(2)     \\ \hline
\textbf{Turning Around} & Discrete(2)       & /                & Discrete(2)     \\ \hline
\textbf{Combined Strategies} & Discrete(3) & \begin{tabular}[c]{@{}c@{}}Box(low=0, \\ high=1, shape=(2,))\end{tabular} & Discrete(14)          \\ \hline
\end{tabular}
\caption{Action Spaces}
\label{tab:action_space}
\end{table*}


\subsubsection{Adversary: Perturbations}
In our bus environment, we parameterize the environment with a perturbation that we treat as an adversary. Instead of the typical adversary, our perturbation adversary just randomly adds a fixed perturbation strength $\alpha$ with probability $0.01$ at each time step on a random bus. The perturbation simply adds $\alpha$ timesteps to the bus's travel time from travelining from station to station. We discretize $\alpha$ from $0$ to $4$. We chose to model an adversary in this way in order to better model real life bus systems where a perturbation would represent the breaking down of a bus that causes an extensive delay in its travel time. 


\subsubsection{Bus Initializations}
In our bus environment, we parameterize a bunching strength that determines the initial bunchedness of the buses. We model the initialization this as a way to model the bunching problems that occurs in bus systems. These bunches create slow downs so we parameterized these initializations to be a part of the curriculum in order to create more instances in training where the buses are bunched and the model is forced to deal with these instances. We do this by defining a $\beta$ which goes from $1$ to $10$ to represent the amount of bunching at the initialization. The buses are initially distributed following Algorithm \ref{alg:bus_init}. The bus initialization is created by first uniformly sampling $\beta$ station indexes from all possible stations, and centering an independent Gaussian with fixed variance on each sampled station. Next, for each bus, one of the created Gaussians is chosen uniformly at random, and the bus's starting station is sampled from this chosen Gaussian. This means that a lower bunching strength value indicates more bunching, and a higher value, in expectation, indicates more of a spread.

\begin{algorithm}
\caption{Bus Initialization Algorithm}\label{alg:bus_init}
\begin{algorithmic}[1]
\State Sample $\beta$ stations $s_1 \cdots s_{\beta}$ uniformly at random with replacement
\For{each station $s_i$ in sampled stations}
    \State $Z_i \gets \mathcal{N}(i, 2.5)$
\EndFor
\For{each bus $j$ in \textbf{NUM BUSES}}
    \State Sample Gaussian $Z_j$ uniformly with replacement from $Z_1, \ldots, Z_n$
    \State Sample station $s_j$ as the start location for bus $b_j$ from Gaussian $Z_j$
\EndFor
\end{algorithmic}
\end{algorithm}


\subsection{Domain Randomization}
\label{sec:dr}
In bus operation deployment, it's crucial for the Reinforcement Learning (RL) algorithm to withstand the unpredictability of actual bus timetables. Factors like unforeseen delays or abrupt increases in passenger numbers can interfere with the efficiency of bus services, leading to complications such as buses clustering together. Consequently, the RL algorithm needs to be designed to cope with these kinds of delays. To bridge the gap between simulation and reality, we incorporated Domain Randomization (DR) in our training process.

The key rationale for DR is to inject variability into the training setting, thereby enhancing the RL agent's ability to adapt to unexpected situations in real-life scenarios. DR facilitates the creation of multiple simulated environments with varied randomized characteristics, aiming to train a model that functions effectively across these scenarios. This model is expected to be versatile in real-world conditions, as the actual system is likely to be a manifestation of the range of training variations.

In DR, the training environment is the primary domain we can fully access, while the actual bus operations represent the target domain for application. We train the model in the primary domain, manipulating a set of \(N\) randomization parameters in this domain \(e_{\xi}\), with the configuration \(\xi\) drawn from a randomization space.

DR models the differences between the primary and target domains as variations in the primary domain. During policy training, we collect episodes from the primary domain \(e_{\xi}\) using randomization, helping the policy familiarize itself with a variety of environments and improve its generalization ability. The policy parameter \(\theta\) is optimized to maximize the average expected reward \(R\) across different configurations:

\[
\theta^*=\arg \max _\theta \mathbb{E}_{\xi \sim \Xi}\left[\mathbb{E}_{\pi_\theta, \tau \sim e_{\xi}}[R(\tau)]\right]
\]

Here, \(\tau_{\xi}\) is a trajectory gathered in the primary domain randomized with \(\xi\). We then apply DR to passenger arrival rates and bus schedules.

Following the methodology of Sánchez-Martínez et al.\cite{sanchez-martinez_real-time_2016}, we set the adjustment factor range from 0 to 2. The chosen arrival rates are designed to result in a peak bus load of 75\% capacity in high-crowding scenarios and 25\% in low-crowding. Similarly, we define three different levels of passenger arrival rates. Without DR, the passenger arrival rate is fixed at \(\lambda\). The demand levels are represented by parameters \(l_1, l_2, l_3\), denoted as \(L=\{l_1, l_2, l_3\}\). We set \(l_1=1.25\) in high-crowding cases, \(l_1=1.0\) in normal scenarios, and \(l_1=0.75\) in low-crowding situations. The adjusted passenger arrival rate with DR is \(\lambda^*\), and demand levels \(l_1, l_2, l_3\) are randomly chosen, so \(\lambda^* = rand(L)\lambda\). This variability equips the RL agent to handle peak and off-peak conditions, and any unplanned spikes in passenger numbers.

Additionally, Gaussian noise is added to the demand levels \(l_1, l_2, l_3\) to further increase randomness. These randomized demand levels are clipped within the initial demand levels array \(L\) to keep them realistic and within set limits. The modified demand levels are then used to determine the passenger arrival rate, influencing the generation of passenger arrival times. This DR approach ensures a wide range of possible scenarios in the generated passenger arrival tables, contributing to a more generalized and adaptable model.

Considering that delays are common in bus operations, we included random delays in our training environment. Training the agent with these random delays aims to make it adaptable to various conditions and ensure smooth operations. The random delay \(rd\) is added every time a bus departs a station, following a uniform distribution: 
$$rd \sim U[\text{min}_{\text{delay}}, \text{max}_{\text{delay}}]$$

We view the learning randomization parameters in DR as a two-level optimization problem. We assume access to real-world bus operations \(e_{real}\) and that the randomization settings (demand level and random delay) are sampled from a distribution parameterized by \(\phi, \xi \sim P_\phi(\xi)\). Our objective is to learn a distribution on which a policy \(\pi_\theta\) can be trained to achieve optimal performance in \(e_{real}\):

$$
\phi^*=\arg \min _\phi \mathcal{L}\left(\pi_{\theta^*(\phi)} ; e_{real}\right) 
$$
$$
\theta^*(\phi)=\arg \min _\theta \mathbb{E}_{\xi \sim P_\phi(\xi)}\left[\mathcal{L}\left(\pi_\theta ; e_{\xi}\right)\right]
$$
where $\mathcal{L}(\pi ; e)$ is the loss function of policy $\pi$ evaluated in the environment $e$.


\subsection{PPO}

Proximal Policy Optimization, more commonly referred to as PPO or the PPO algorithm, is a widely used optimization algorithm introduced by \cite{schulman2017proximal} in 2017 for training agents in sequential decision making tasks. It's similar to the well known REINFORCE algorithm, and is another policy gradient method. We decide to use this algorithm in our experiments due to its tendency for high stability and sample efficiency.


\subsection{Setter-Based Curriculum Learning}
\paragraph{Setter Model} Based on the framework proposed by Sheng et al.\cite{DBLP:journals/corr/abs-1909-12892}, our setter  model creates 'lessons' which update the environment in which the model is trained on. To define the setter, we first must define its input parameters. The input are the previous $3$ lessons $\phi$ and the previous $3$ rewards. We define a lesson as a part of a curriculum that contains (S, $\alpha$, $\beta$) to parameterize the environment. The first part of the lesson $\phi$ is the $S$ which is an index over the action space which corresponds to a specific action mask which is a combination of holding values, skipping, and turning (reference Appendix Figure \ref{tab:action_space} for more specific details). Though not every one of the possible $63$ combinations of available actions was given to the model, the action mask is discretized using a representative group of 15 combinations. The second part of the lesson $\phi$ is \textit{adversary strength} ($\alpha$), which we define as the expected total time of delays caused by random perturbations affecting buses. The third part of the lesson $\phi$ is \textit{bunching strength} ($\beta$), which we define as the distribution of buses among the stations when the environment is initially created, as described in an earlier section. The final input is the previous rewards corresponding to after training on these previous lessons. Similar to the REINFORCE algorithm, we use mean log probabilities to estimate the gradient over the action space. The setter model loss is defined as follows (where $\pi$ represents the probability distribution over the state space):
\[\mathcal{L}_{setter} = -r_{[t-3:t]} \cdot (\log\pi_S + \log\pi_\alpha + \log\pi_\beta)\]
As seen in the equation above, the mean reward is weighted by the log probabilities of each of the three other components of the setter model loss ($S, \alpha, \beta)$, and this is used to perform gradient descent on the model to update its parameters. 
\paragraph{Environment Update Algorithm}
Setter-based curriculum learning makes use of the setter model described in the previous section to adjust environment variables for the agent as it learns the optimal policy. As mentioned earlier, it uses its input variable to compute ''lessons" in the form of adjusted environments where the agent can learn in. We use Algorithm \ref{tab:sb_curr} to update the environment using the setter. At each update step in Algorithm \ref{tab:sb_curr}, the model uses rewards, available actions, adversary strength, and bunching strength from the previous three time steps to compute the curriculum for the next time step. The rewards from this time step are then passed back into the model, which updates its parameters and potentially generates a different curriculum for the next curriculum update step.

\begin{algorithm}
  \caption{Setter-Based Curriculum Learning} \label{tab:sb_curr}
  \textbf{Initialize:} Environment $\mathcal{E}$, rollout buffer $D$, agent’s policy $\pi_{\theta}(a | s)$ and corresponding parameter $\theta$ \\
  \textbf{Initialize:} Setter model $\tau$, step size $\eta$, available actions $S$, adversary strength $\alpha$, bunching strength $\beta$, global steps $T_{\text{total}}$

  \begin{algorithmic}[1]
    \For{$i$ in $N_{\text{iter}}$}
      \State $\theta_i \gets \theta_{i-1}$
      \State Compute lesson $\phi$ from $\tau$ such that $\phi = \tau(r_{t-3:t}, S_{t-4:t-1}, \alpha_{t-3:t}, \beta_{t-3:t})$
      \State Update environment according to $\phi$
      \For{$t$ in $T_{\text{trajectory}}$}
        \State Sample action $a_t \sim \pi_{\theta}(a | s)$ in environment
        \State Get the next state $s_{t+1}$ by taking $A_{\text{adv}}$, and store $(s_t, A_{\text{adv}}, r_t, s_{t+1})$ in $D$
      \EndFor
      \State Update $\tau$ using $r_{t}, S_{t}, \alpha_{t}, \beta_{t}$
      \State Optimize parameter $\theta_i$ using $D$
    \EndFor
  \end{algorithmic}
\end{algorithm}


\section{Experiments}
\subsection{Experimental Setting}

\begin{figure*}[h]
    \centering
\includegraphics[width=1\textwidth]{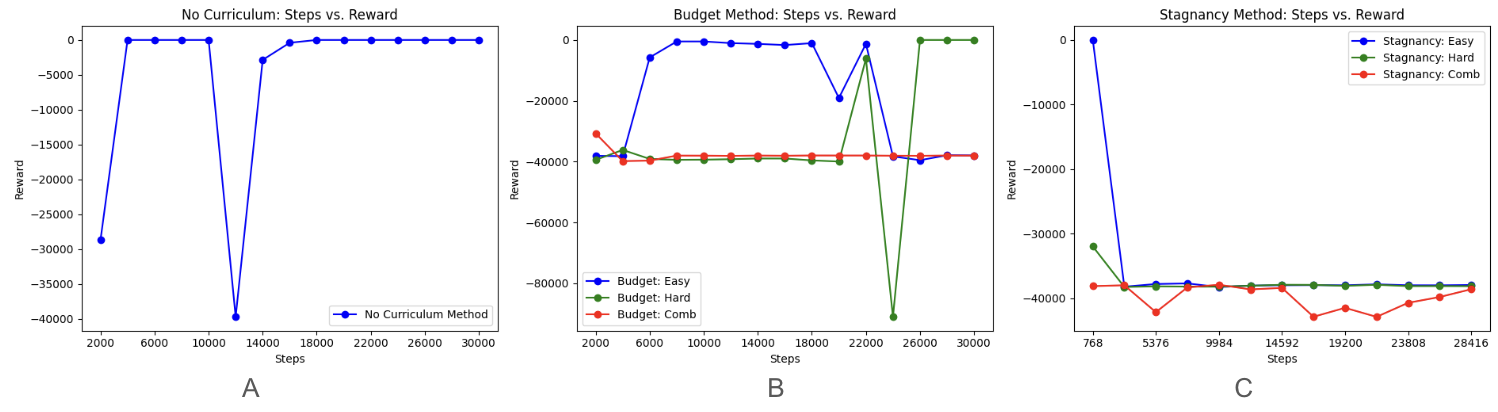}
    \captionof{figure}{The three baseline methods used in our experiments}
    \label{fig:nc_b_s}
\end{figure*}


\paragraph{Bus Environment}
We run all our experiments using the same Bus Environment environment described in Sec.\ref{sec:bus_env_1} and use the same parameters for all of our runs. We use 10 bus stations, 14 buses, bus capacity of 60, distance between stations of 1 km. 

\paragraph{Baselines} We use the following $3$ methods as a baseline to compare our setter-based curriculum learning model. We trained each curriculum learning model on an easy-first curriculum, hard-first curriculum, and combination curriculum and take the best as the baseline for each respective curriculum learning method. For our experiments we define the we define the easy-first curriculum as seen in Appendix Figure \ref{tab:easy_first_config}, the hard-first curriculum as seen in Appendix Figure \ref{tab:hard_first_config}, and the combination curriculum as seen in Appendix Figure \ref{tab:various_combinations_two}. The baseline curriculum learning methods are as follow: 
\begin{itemize}
    \item \textbf{No Curriculum Learning:} Under this baseline the model has access to all the actions during all time steps and the bus placement initialization is the standard intialization where all buses start in the same station and gradually leave one by one. In addition, there is no random perturbations. 

    \item \textbf{Budget Based Curriculum:} This framework is inspired by the approach used by Sheng et al. \cite{sheng2022curriculum}. Unlike their paper, however, our curriculum does not monotonically increase in difficulty, in fact in some experiments, it is set to decrease. Essentially, a total "budget" is given throughout all training steps, and each environment state is preset in advance. The curriculum used for a given timestep is defined as follows:
    \[\phi_t = b_m \text{ if } t \in [t_m, t + T_{b_m})\]
    The curriculum is discretely defined in different ways for different portions of the total training steps.

    \item \textbf{Stagnancy Based Curriculum:} The stagnancy framework is based on the idea that if the reward is stalled for a certain period of time, then the environment should be changed to potentially induce an increase in reward. Essentially, if reward does not increase by a certain amount within a certain number of timesteps, the environment will either be made easier or more difficult depending on a preset curriculum. This framework was inspired by curriculum adversarial training of  \cite{cai2018curriculum}, but adapted to suit our environment.
\end{itemize}

\subsection{Results}


\subsubsection{Baselines}
\paragraph{No Curriculum Method} The rewards for the No Curriculum Method is in Figure \ref{fig:nc_b_s}\textcolor{red}{.A} which works quite well as it is able to quickly reach high rewards and is fairly stable. 

\paragraph{Budget Based Curriculum Method}
We compare the performance of our budget based curriculum on the $3$ different fixed curriculums, easy-first, hard-first, and combination. We compare these methods and use the best for comparison as the baseline model for the budget based curriculum. From Figure \ref{fig:nc_b_s}\textcolor{red}{.B} we can tell that an easy-first curriculum works best based on stability and reward, thus we use the easy-first curriculum for the Budget Based method as the baseline. Exact experiment input can be seen in Appendix tables \ref{tab:easy_first_config}, \ref{tab:hard_first_config}, and \ref{tab:various_combinations_two} 

\paragraph{Stagnant Based Curriculum}
We compare the performance of our stagnancy based curriculum on the $3$ different fixed curriculums, easy-first, hard-first, and combination. We compare these methods and use the best for comparison as the baseline model for the stagnancy based curriculum. From Figure \ref{fig:nc_b_s}\textcolor{red}{.C} we can tell that an easy-first curriculum works best thus we use the easy-first curriculum for the Stagnant Based method as the baseline.

\subsubsection{Overall Comparison}
\begin{figure}[h]
    \centering
    \includegraphics[scale=0.5]{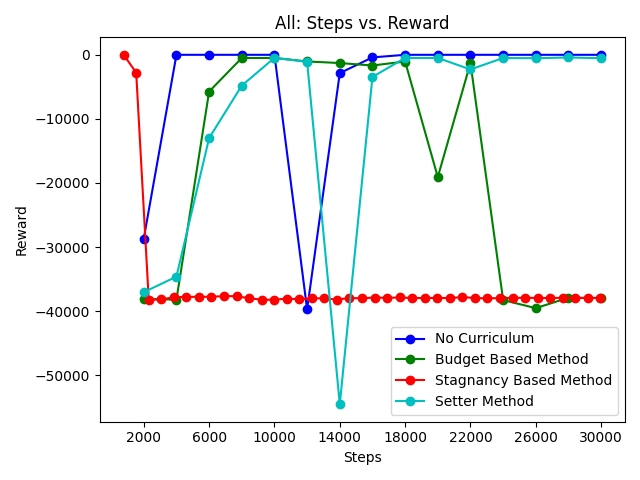}
    \caption{Comparing the baseline methods to our Setter Curriculum Method}
    \label{fig:compare}
\end{figure}

The results for the overall comparison of the various frameworks can be seen in Figure \ref{fig:compare}. The setter model outperforms the budget method and the stagnation methods seen above, but seems to slightly underperform when compared to the no curriculum method. This seems to suggest that it has potential, but in its current state could require more finetuning. The setter has potential as it does not require as much domain specific knowledge and time designing curriculum with various lessons yet is still able to outperform these methods. We also suspect that the model suffers from high variance as we use the REINFORCE algorithm with sampling to get the rewards that the setter model uses a signal. However, long evaluation times limited the number of trajectories we could use and the length of these trajectories, thus forcing us to keep these relatively short and thus even higher variance.

\subsection{Analysis}
\subsubsection{Setter Curriculum}
In this section, we analyze the the actually Curriculum that our Setter Model generates during training. We examine the Setter Model's Lessons $\phi$ which consist of the Action Space (S), Perturbation Strength ($\alpha$), and Bunching Strength ($\beta$) values.

\begin{figure}[h]
    \centering
    \includegraphics[scale=0.5]{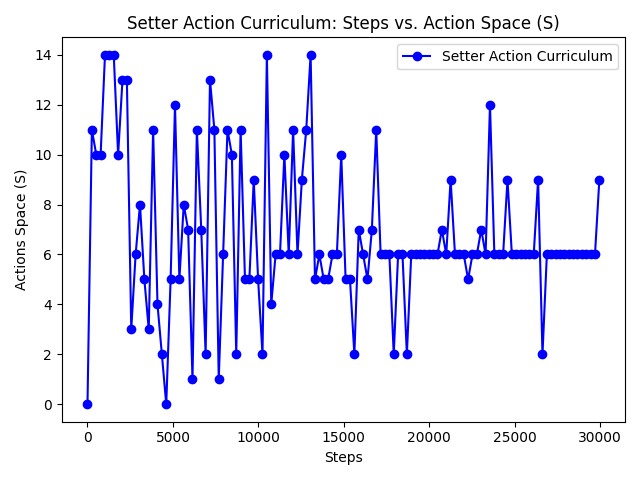}
    \caption{Visualizing the Action Space ($S$) of the different lessons $\phi$ from the Setter Model}
    \label{fig:curr_act}
\end{figure}

\paragraph{Action Space (S)} As seen in Figure \ref{fig:curr_act}, the setter model experimented with both high and low action spaces earlier on, then started to converge around a value of $6$ available actions at around $17000$ steps. This seems to imply that $6$ is an optimal value for model performance with regard to the actions that the agent is allowed to take. This hints that an effective curriculum method experiments with different lessons and will thus use which ever lesson works best for learning. It is also important to note that even while the model is primarily outputting $S=6$ the model will still occasionally experiment with other values, which could be that as the reward plateaus the model tries different $S$ values. 

\begin{figure}[h]
    \centering
    \includegraphics[scale=0.5]{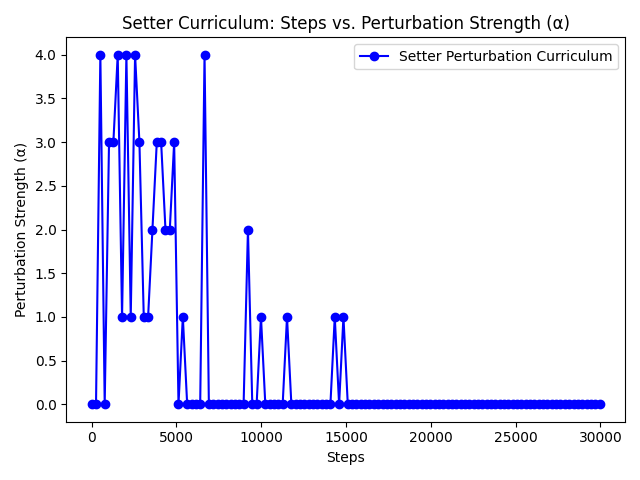}
    \caption{Visualizing the Perturbation Strength ($\alpha$) of the different lessons $\phi$ from the Setter Model}
    \label{fig:curr_adv}
\end{figure}

\paragraph{Perturbation Strength ($\alpha$)} As seen in Figure \ref{fig:curr_adv}, the perturbation strength follows a similar pattern to the Action Space Curriculum with more exploration towards the beginning and the converging to a perturbation strength of $0$. At the beginning the model seems to explore a lot more with the higher values and almost never explores $0$ but as the $\alpha$ converges to 0 there is less exploration above 0 and explorations only go to 1. This could hint that while the reward is also converging and the model is trying to reach higher rewards, the perturbation strength as we have defined is not as key of a part of the curriculum. In addition, the perturbation strengths also converge to $0$ at around $5000$ steps which could hint that the model was able to more quickly learn that $\alpha$ is not an effective way of achieving more performance. 

\begin{figure}[h]
    \centering
    \includegraphics[scale=0.5]{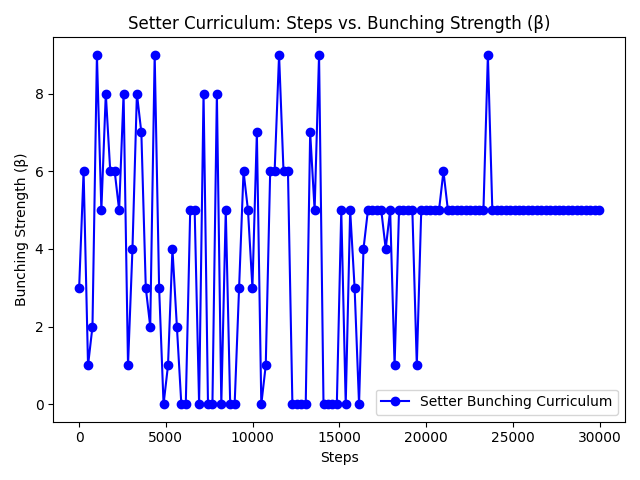}
    \caption{Visualizing the Bunching Strength ($\beta$) of the different lessons $\phi$ from the Setter Model}
    \label{fig:curr_bun}
\end{figure}

\paragraph{Bunching Strength ($\beta$)} As seen in Figure \ref{fig:curr_bun}, the bunching strength follows a similar pattern to both the Action Space (S) and the Perturbation Strength ($\alpha$) where the beginning is a lot of exploration and variation in $\beta$ values but begins to converge around $20000$ steps. This could indicate that bunching strength $\alpha$ can add a fair amount of variation to the training and is thus used heavily while the reward has not converged. While the $\beta$ values converge to $5$ there is not as many spikes as in the action space which could indicate that it is less effective in increasing reward as a curriculum.

\subsection{Ablation Study}
\subsubsection{Contribution from Domain Randomization}
In this section we ablate between using Domain Randomization \cite{tobin2017domain} as described in Section \ref{sec:dr}. In this experiment, we use no curriculum learning and all other standard settings. We find that without Domain Randomization the model stagnates at a considerably lower reward. This demonstrates that using Domain Randomization vastly improves the performance and this is likely due to the model benefiting from experiencing a larger variety of training situations which are more representative of the variety of situations that can arise during test time. In addition, in Bus Systems where things like bus bunching can form from these small random delays in the buses, Domain Randomization is a core part of the methodology to appropriately model these situations and create a robust bus system that can handle them.

\begin{figure}[h]
    \centering
    \includegraphics[scale=0.5]{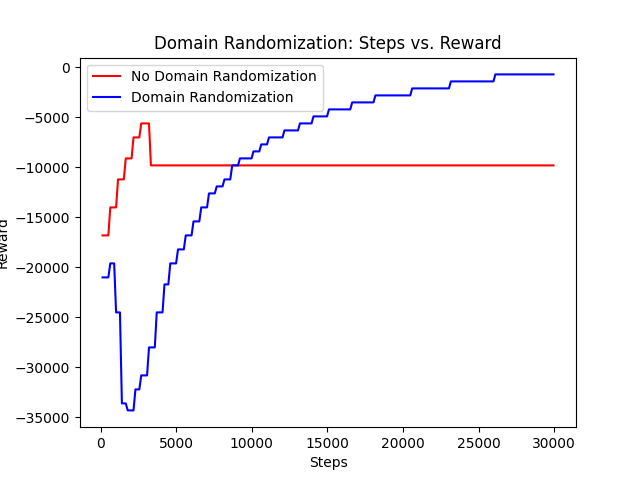}
    \caption{Ablating Domain Randomization}
    \label{fig:dr_ab}
\end{figure}

\subsubsection{Setter Curriculum}
\begin{figure}[h]
    \centering
    \includegraphics[scale=0.5]{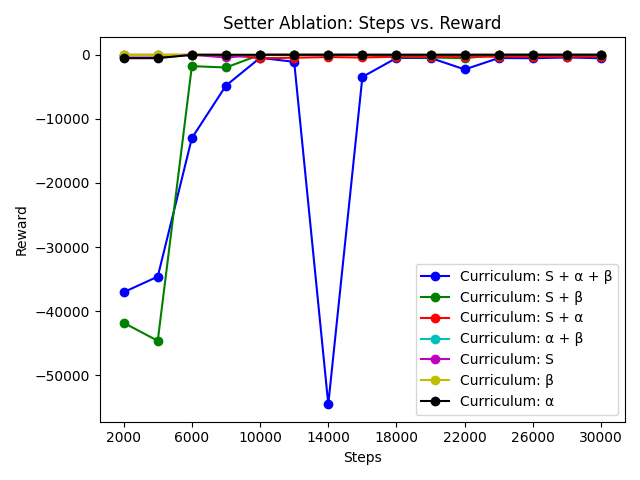}
    \caption{Ablating Setter Curriculum Lessons $\phi$}
    \label{fig:setter_ab}
\end{figure}

In this ablation study we ablate over the lessons $\phi$ and whether the model has control over the action space (S), perturbation strength ($\alpha$), and bunching strength ($\beta$). In this experiment we experiment with all the variations of different Curriculum possibilities as seen in Figure \ref{fig:setter_ab}. We find that experiments where the Setter only affects $\alpha$, $\beta$, or S the performance is better which could signify that having the setter create lessons $\phi$ based on all 3 of even combinations overloads the model and the model is unable to do it as effectively. This could signify that our Setter model does not have enough capacity to handle all of these options and is able to perform better. This is similar when looking at the combinations as $S + \alpha$ and $\alpha + \beta$ are able to perform better than $S + \alpha + \beta$ which hints the model could be overloaded without enough capacity. 

\section{Conclusion}
Throughout this paper, we described a new method for generating dynamic curricula throughout the training process of a reinforcement learning agent. The setter model takes in various environmental parameters, as well as previous rewards, and attempts to learn the optimal next environment to induce faster learning for the agent. We demonstrated that this approach has potential, since it outperformed two of the three baselines that we provided. There is still much more room for development, however, potentially in different environments and using different architectures. One approach in particular could be to try a more complex model or loss function. Since the baseline model with no curriculum learning already performs well, perhaps a more difficult environment overall could be more suitable for curriculum learning based frameworks.

\section{Contribution}

\paragraph{Group Contribution} Danny and Avi are enrolled in CS285. Yuhan was originally on the waitlist while forming the group but ended up just auditing the class. Yuhan primarily implemented the basic bus environment and the domain randomization. Avi and Danny primarily worked together to implement the setter and the necessary framework and ran experiments. The writing of the paper was generally split evenly between everyone.


\newpage~\newpage
{\small
\bibliographystyle{ieee_fullname}
\bibliography{egbib}
}

~\newpage~\newpage
\section{Appendix}
\begin{table}[h]
  \centering
  \begin{tabular}{|c|c|c|c|}
    \hline
    \textbf{Action Space (S)} & \textbf{Holding} & \textbf{Skipping} & \textbf{Turning} \\
    \hline
    0 & 0 holding & Yes & Yes \\
    1 & 0 holding & Yes & No \\
    2 & 0 holding & No & Yes \\
    3 & 0-40 holding & Yes & No \\
    4 & 0-40 holding & Yes & Yes \\
    5 & 0-40 holding & No & No \\
    6 & 0-40 holding & No & Yes \\
    7 & 0-80 holding & Yes & No \\
    8 & 0-80 holding & Yes & Yes \\
    9 & 0-80 holding & No & No \\
    10 & 0-80 holding & No & Yes \\
    11 & 0-120 holding & Yes & No \\
    12 & 0-120 holding & Yes & Yes \\
    13 & 0-120 holding & No & No \\
    14 & 0-120 holding & No & Yes \\
    \hline
  \end{tabular}
  \caption{Action Space (S)}
  \label{tab:action_space}
\end{table}

\begin{table}[h]
  \centering
  \begin{tabular}{|c|c|c|c|c}
    \hline
    \textbf{Difficulty Level} & \textbf{action\_mask\_levels} & \textbf{perturbation\_levels} & {bunching\_levels} & 
    \textbf{difficulty\_thresholds} \\
    \hline
    1 & 0 & 0 & 10 & 0.01\\
    2 & 0,1 & 0 & 10 & 0.01\\
    3 & 0,1,2 & 0 & 9 & 0.08\\
    4 & 0,1,2,3 & 1 & 9 & 0.04\\
    5 & 0,1,2,3,4 & 1 & 8 & 0.04\\
    6 & 0,1,2,3,4,5 & 1 & 8 & 0.04\\
    7 & 0,1,2,3,4,5,6 & 2 & 7 & 0.04\\
    8 & 0,1,2,3,4,5,6,7 & 2 & 7 & 0.04\\
    9 & 0,1,2,3,4,5,6,7,8 & 2 & 6 & 0.04\\
    10 & 0,1,2,3,4,5,6,7,8,9 & 3 & 6 & 0.04\\
    11 & 0,1,2,3,4,5,6,7,8,9,10 & 3 & 5 & 0.04\\
    12 & 0,1,2,3,4,5,6,7,8,9,10,11 & 3 & 5 & 0.04\\
    13 & 0,1,2,3,4,5,6,7,8,9,10,11,12 & 4 & 4 & 0.04\\
    14 & 0,1,2,3,4,5,6,7,8,9,10,11,13 & 4 & 4 & 0.1\\
    15 & 0,1,2,3,4,5,6,7,8,9,10,11,12,13 & 4 & 3 & 0.1\\
    16 & 0,1,2,3,4,5,6,7,8,9,10,11,12,13 & 4 & 3 & 0.3\\
    \hline
  \end{tabular}
  \caption{Easy First Configuration}
  \label{tab:easy_first_config}
\end{table}

\newpage~\newpage
\begin{table}[h]
  \centering
  \begin{tabular}{|c|c|c|c|c|}
    \hline
    \textbf{Difficulty Level} & \textbf{action\_mask\_levels} & \textbf{perturbation\_levels} & \textbf{bunching\_levels} & \textbf{difficulty\_thresholds} \\
    \hline
    1 & 0 & 4 & 3 & 0.01 \\
    2 & 0,1,2,3,4,5,6,7,8,9,10,11,12,13 & 4 & 3 & 0.19 \\
    3 & 0,1,2,3,4,5,6,7,8,9,10,11,13 & 4 & 4 & 0.038 \\
    4 & 0,1,2,3,4,5,6,7,8,9,10,11,12 & 4 & 4 & 0.038 \\
    5 & 0,1,2,3,4,5,6,7,8,9,10,11 & 3 & 5 & 0.038 \\
    6 & 0,1,2,3,4,5,6,7,8,9,10 & 3 & 5 & 0.038 \\
    7 & 0,1,2,3,4,5,6,7,8,9  & 3 & 6 & 0.038 \\
    8 & 0,1,2,3,4,5,6,7,8 & 2 & 6 & 0.038 \\
    9 & 0,1,2,3,4,5,6,7 & 2 & 7 & 0.038 \\
    10 & 0,1,2,3,4,5,6 & 2 & 7 & 0.038 \\
    11 & 0,1,2,3,4,5 & 1 & 8 & 0.038 \\
    12 & 0,1,2,3,4 & 1 & 8 & 0.038 \\
    13 & 0,1,2,3 & 1 & 9 & 0.038 \\
    14 & 0,1,2 & 0 & 9 & 0.038 \\
    15 & 0,1 & 0 & 10 & 0.038 \\
    16 & 0,1,2,3,4,5,6,7,8,9,10,11,12,13 & 2 & 10 & 0.3 \\
    \hline
  \end{tabular}
  \caption{Hard First Configuration}
  \label{tab:hard_first_config}
\end{table}

\begin{table}[h]
  \centering
  \begin{tabular}{|c|c|c|c|c|}
    \hline
    \textbf{Difficulty Level} & \textbf{action\_mask\_levels} & \textbf{perturbation\_levels} & \textbf{bunching\_levels} & \textbf{difficulty\_thresholds} \\
    \hline
    1 & 0 & 2 & 1 & 0.01 \\
    2 & 0,1,2,3 & 2 & 1 & 0.06 \\
    3 & 0,1,2,3,4,5,6 & 2 & 1 & 0.07 \\
    4 & 0,1,2,3,4,5,6,7,8,9,10,11 & 2 & 1 & 0.07 \\
    5 & 12 & 2 & 1 & 0.07 \\
    6 & 13 & 2 & 1 & 0.07 \\
    7 & 0,1,2,3,4,5,6,7,8,9,10,11,12 & 2 & 1 & 0.07 \\
    8 & 0,1,2,3,4,5,6,7,8,9,10,11,13 & 2 & 1 & 0.07 \\
    9 & 12,13 & 2 & 1 & 0.07 \\
    10 & 0,1,2,3,4,12 & 2 & 1 & 0.07 \\
    11 & 0,1,2,3,4,13 & 2 & 1 & 0.07 \\
    12 & 0,1,2,3,4,5,6,7,8,9,10,11,12,13 & 2 & 1 & 0.07 \\
    \hline
  \end{tabular}
  \caption{Various Combinations Configuration}
  \label{tab:various_combinations_two}
\end{table}

\end{document}